# Multi LoRA Meets Vision: Merging multiple adapters to create a multi task model

Ege Kesim
Huawei R&D Turkiye
ege.kesim1@huawei.com

Selahattin Serdar Helli
Huawei R&D Turkiye
serdar.helli2@huawei-partners.com

*Abstract*— Parameter efficient finetuning (PEFT) methods are widely used in LLMs and generative models in computer vision. Especially one can use multiple of these during inference to change the behavior of the base model. In this paper we investigated whether multiple LoRA adapters trained on computer vision tasks can be merged together and used during inference without loss in performance. By achieving this, multitask models can be created just by merging different LoRAs. Merging these will reduce inference time and it will not require any additional retraining. We have trained adapters on six different tasks and evaluated their performance when they are merged together. For comparison we used a model with a frozen backbone and finetuned its head. Our results show that even with simple merging techniques creating a multitask model by merging adapters is achievable by slightly loosing performance in some cases. In our experiments we merged up to three adapters together. Depending on the task and the similarity of the data adapters were trained on, merges can outperform head finetuning. We have observed that LoRAs trained with dissimilar datasets tend to perform better compared to model trained on similar datasets.

## I. INTRODUCTION

Finetuning large pretrained models on different tasks and domains has been popular for a long time. However as these pretrained models get larger and larger it becomes computationally much more expensive to finetune them. Parameter efficient finetuning (PEFT) methods became popular recently as they enable finetuning models with less resources. Some popular PEFT methods are LoRA [1], DoRA[2], O-LoRA[3], Q-LoRA [4] and MultiLoRA [5].

Some of these PEFT methods such as LoRA enable only storing much smaller weight files to be added on top of the base model. In a case where there are multiple finetuned versions of a large model, by using LoRA it is enough to only store the small weight files for each task rather than the full model. Multiple LoRA adapters can also be merged in order to add new task capabilities to a model without causing catastrophic forgetting. Due to these reasons, LoRA finetuning is heavily used in in large language models (LLMs). They are especially used in LLMs to add new tasks by merging with a LoRA or efficiently finetune the model with new data to update the knowledge. Using multiple LoRA adapters is much more popular in LLMs compared to computer vision [3,6,7,8].

In computer vision multiple LoRA merging is almost only used in generative tasks [9,10,11,12,13,14]. Each LoRA usually correspond to a style or character. By merging these, new styles and characters can be added to base model. This way multiple concepts and styles can be used at once and capabilities of the model can be extended without any additional retraining for merging.

Other than generative models LoRA finetuning is also used in computer vision for various tasks such as segmentation [16], classification [15,17] and object detection [18]. However, as it was mentioned earlier using multiple LoRAs is almost only used in generative tasks.

Using multiple LoRAs in vision has multiple benefits:

*1) Without any multitask training multiple single task models can be merged to create a multitask model.*
*2) Computation cost can be reduced for multiple task cases since there will be a shared backbone which is the bottleneck of the network due to its size.*
*3) A network can be updated any time. New tasks can be added and old ones can be removed from the network just by removing the LoRA.*
*4) In the need of retraining a task, instead of retraining the whole model for all tasks it will be enough to just train the LoRA of that task.*

Using multiple LoRA adapters requires each LoRA to work without affecting each other. In this work we explored whether multiple LoRA models sharing the same base model can be merged together to create a multi task system. We selected four classification and two regression tasks on five different datasets to explore how adapters of these models interact. One of these datasets consists of satellite images of earth, another dataset consists of galaxy images and the rest of them are made of face images. We have purposefully selected these datasets to see how datasets with similar and different content will affect the models when they are merged. We have also compared regression and classification tasks to see if the task type is affecting due to the pretraining of the backbone model. To see the how effective to use LoRA we compared using LoRA with different ranks and, finetuned model with a frozen backbone as a baseline. By merging multiple heads with the frozen backbone, a multitask model can be created easily without performance loss due to merging. However only finetuning the head may not give enough performance for every task due to low number of parameters that are trained.

In our experiments we observed that merging multiple LoRA adapters result in performance drop. The amount of the drop is determined by the adapter combination and this selection can make a huge difference in the performance. We observed that among all adapters we have trained, two of the adapters preserve performance when combined with almost every other adapter. Depending on the combined tasks and datasets, the merged model can outperform the scenario where only the heads are finetuned.

## II. DATASETS

In our experiments we have used five different datasets for six different tasks. These datasets were chosen so that the data is not similar to the data that the base model was pretrained with. We chose four face related datasets expecting the adapters to learn similar features and two dissimilar datasets. In order to have similar dataset sizes we down sampled each dataset to a size around 10-17k samples.

### A. FireRisk

FireRisk [19] is a dataset used in remote sensing. It consists of satellite images each labeled among 7 classes assessing the fire risk. There are 91k images but the distribution of data among these classes is highly imbalanced. As it was mentioned before to get each dataset to similar size we have down sampled each dataset. In FireRisk we have reduced the number of classes to 3 classes: non-burnable, high, and low. To reduce the classes, we combined water and non-burnable, high, very high and moderate and lastly low and very low together to 3 bins. After combining these we randomly sampled data from each bin to reduce the amount of data.

### B. UTKFace

UTKFace [20] consists of 23k images. Each image is annotated for age, gender and ethnicity. We used this dataset for two different tasks to predict age which are classification and regression. For classification we divided the dataset to 6 classes. The classes are ages of 0-3, 4-16, 17-30, 31-45, 46-59 and 60+. After appending each image to one of these bins, the dataset was down sampled to 11k images. Each class is down sampled to 2k images except 0-3 class due to low number of images in that class. The same down sampled dataset was used for regression tasks without any change. We have used Dlib [21] face detector to detect and crop face in images.

### C. Expw

Expression in the Wild (Expw) [22,23] is a dataset containing 91k face images. Each image contains a person expressing one of the 7 basic emotion categories. We have reduced the dataset to 12k images.

### D. WFLW

WFLW [24] is a popular dataset used in facial landmark detection. It consists of 7500 train and 2500 test images. In each image landmarks are represented with 98 2D points. Compared to the other datasets the diversity of images is high in terms of expression, poses etc. Also, the test set images are annotated for these six categories.

### E. Galaxy10 DECals

Galaxy10 DECals [25] is a dataset containing galaxy images with 10 classes. The total number of images is around 17k. The dataset is created using Galaxy Zoo [26] and DESI Legacy Imaging Surveys [27,28,29]. We removed the class "Cigar Shaped Smooth Galaxies" which only had 334 samples in total. This class had much lower sample size compared to all other classes. The resulting data consists of 9 classes.

## III. METHOD

We employ a pre-trained Vision Transformer (ViT) [30], as the backbone of our model. Additionally, we compare the efficacy of using LoRA against a baseline of fine-tuning a model with a frozen backbone, and the multi-merged LoRA models. For merging multiple LoRA adapters, concatenation merging method was used. Although we have not reported the results we also ran experiments for linear merging and observed similar results.

We utilize the ViT model as the backbone. The model's pooler output provides a 768-dimensional feature vector. Following the backbone a two-layer Multi-Layer Perceptron (MLP) is applied to each task. These MLPs consist of a 512-dimensional hidden layer with ReLU activation, followed by an output layer sized according to the task-specific requirements. LoRA is applied to the key and value vectors of the backbone.

For classification tasks we employ the cross-entropy loss which are fire risk classification, galaxy classification, emotion recognition and age group classification. For the task of face age regression and landmark detection we utilize L1 loss. Furthermore we report the Normalized Mean Error (NME) score for facial landmark detection task. NME is a common metric for evaluating the accuracy of landmark detection, calculated as the mean error between the predicted and true landmark positions, normalized by an inter-ocular distance to account for scale variations across different faces. In our experiments we used Adam optimizer.

All datasets were divided into training, test, and validation sets. The models were trained until they converged. We report the performance of LoRA adapters on six different tasks: fire risk classification, age classification (UTKFace), emotion recognition (Expw), galaxy classification (Galaxy10 DECals), age regression (UTKFace), and landmark detection (WFLW). These tasks are evaluated based on accuracy (Acc), macro F1 score, root mean square error (RMSE), and normalized mean error (NME) metrics, providing a comprehensive understanding of each model's effectiveness.

In our experiments we compared 16 and 64 rank LoRA finetuning and only finetuning the head of a model with frozen backbone. Performance comparison of these models can be found in Table 1. Although there is no conclusion in which rank is better compared to other, in all scenarios LoRA performs better compared to the baseline. In landmark detection and galaxy classification the baseline method performs much worst meaning the backbone needs some level of finetuning. It seems that in most cases the baseline has comparable performance with LoRAs. We believe that since landmark detection needs pixel coordinates the information extracted from the backbone might need adjustments. Galaxy10 DECals dataset may also need additional changes in the backbone due to its data source difference with the data the pretrained model was trained on.

TABLE I. PERFORMANCE COMPARISON OF DIFFERENT LORA RANKS AND ONLY HEAD FINETUNING

|  | FireRisk | | UTK Face (Classification) | | Expw | | UTK Face (Regression) | WFLW | Galaxy10 DECals | |
| --- | --- | --- | --- | --- | --- | --- | --- | --- | --- | --- |
|  | Acc (↑) | F1 (↑) | Acc (↑) | F1 (↑) | Acc (↑) | F1 (↑) | RMSE (↓) | NME (↓) | Acc (↑) | F1 (↑) |
| LORA-16 | 73.5 | 73.4 | **73.4** | **73.7** | **48.2** | **46.0** | 0.940 | 8.483 | 79.1 | 76.3 |
| LORA-64 | **76.7** | **76.7** | 72.4 | 73.2 | 45.6 | 42.3 | **0.868** | **8.231** | **81.2** | **79.8** |
| Fine tuning head | 75.5 | 75.3 | 72.0 | 72.6 | 42.0 | 40.3 | 0.911 | 25.560 | 64.9 | 63.1 |

TABLE II. PERFORMANCE OF MERGED ADAPTERS ON AGE CLASSIFICATION TASK

| Merged adapters | Age Classification (F1 Score) | |
| --- | --- | --- |
|  | *Lora Rank 16* | *Lora Rank 64* |
| AC+E | 71.6 | 67.6 |
| AC+F | 72.7 | 73.1 |
| AC+L | 7.6 | 9.4 |
| AC+AR | 21.3 | 32.2 |
| AC+G | **73.7** | **73.6** |

TABLE III. PERFORMANCE OF MERGED ADAPTERS ON EMOTION CLASSIFICATION TASK

| Merged adapters | Emotion Classification (F1 Score) | |
| --- | --- | --- |
|  | *Lora Rank 16* | *Lora Rank 64* |
| E+AC | 44.9 | 41.0 |
| E+F | 45.4 | **42.0** |
| E+L | 6.2 | 6.9 |
| E+AR | 29.9 | 39.1 |
| E+G | **47.2** | 41.7 |

TABLE IV. PERFORMANCE OF MERGED ADAPTERS ON FIRE RISK CLASSIFICATION TASK

| Merged adapters | Fire risk Classification (F1 Score) | |
| --- | --- | --- |
|  | *Lora Rank 16* | *Lora Rank 64* |
| F+E | 74.0 | 76.5 |
| F+AC | **74.3** | **76.6** |
| F+L | 59.0 | 65.1 |
| F+AR | 72.9 | 75.9 |
| F+G | 74.0 | 75.9 |

TABLE V. PERFORMANCE OF MERGED ADAPTERS ON LANDMARK DETECTION TASK

| Merged adapters | Landmark Detection (NME) | |
| --- | --- | --- |
|  | *Lora Rank 16* | *Lora Rank 64* |
| L+E | 9.455 | 11.049 |
| L+F | 8.739 | **8.256** |
| L+AC | 9.744 | 9.944 |
| L+AR | 18.716 | 11.787 |
| L+G | **8.550** | 8.436 |

TABLE VI. PERFORMANCE OF MERGED ADAPTERS ON AGE REGRESSION TASK

| Merged adapters | Age Regression (RMSE) | |
| --- | --- | --- |
|  | *Lora Rank 16* | *Lora Rank 64* |
| AR+E | 0.990 | 1.371 |
| AR+F | **0.952** | **0.861** |
| AR+L | 2.885 | 2.663 |
| AR+AC | 1.092 | 1.268 |
| AR+G | 0.955 | 0.877 |

TABLE VII. PERFORMANCE OF MERGED ADAPTERS ON GALAXY CLASSIFICATION TASK

| Merged adapters | Galaxy Classification (F1 Score) | |
| --- | --- | --- |
|  | *Lora Rank 16* | *Lora Rank 64* |
| G+AC | 76.7 | 79.9 |
| G+E | **76.9** | **80.3** |
| G+F | 74.0 | 77.9 |
| G+L | 58.3 | 54.6 |
| G+AR | 75.8 | 80.0 |

Tables 2-7 shows the performance of the merged adapters on each dataset. Each task is represented with their capitals: age classification (AC), age regression (AR), emotion recognition (E), fire risk classification (F), landmark detection (L) and galaxy classification (G). In all cases merging adapters decrease performance with varying levels.

TABLE VIII.  PERFORMANCE OF THREE ADAPTER MERGE

| Dataset | AC - F1 (↑) | | E - F1 (↑) | | AR - RMSE (↓) | | L - NME (↓) | |
|---|---|---|---|---|---|---|---|---|
| | Rank 16 | Rank 64 | Rank 16 | Rank 64 | Rank 16 | Rank 64 | Rank 16 | Rank 64 |
| FireRisk | 73.7 | 76.5 | 73.7 | 75.9 | 73.4 | 76.2 | 63.4 | 65.0 |
| Galaxy10 DECals | 73.8 | 77.7 | 73.3 | 77.9 | 72.7 | 76.8 | 57.8 | 45.6 |
| UTK Face (Classification) | 73.8 | 72.7 | - | - | - | - | - | - |
| Expw | - | - | 46.7 | 41.0 | - | - | - | - |
| UTK Face (Regression) | - | - | - | - | 0.959 | 0.868 | - | - |
| WFLW | - | - | - | - | - | - | 8.855 | 8.431 |

## A. Effect of merging on each task

In this section we examine how each adapters performance change on their own task when merged with other adapters. Fire risk and galaxy classification adapters are resilient to merging except the merge with the landmark adapter. From Tables 4,7 it can be seen that these adapters have only a slight decrease in their performance when merged with other models. However even with this slight decrease they still outperform the baseline.

Landmark adapter lose performance in each case. It works best with galaxy and fire risk adapter which are non face containing tasks. The adapters that have dataset with similar content creates a performance drop compared to galaxy and fire risk adapters. However landmark model still outperforms baseline as the baseline performs poorly on this task. Overall landmark adapter lose performance with other merges however it still performs well.

Age regression adapter acts similar to landmark adapter. It performs best when merged with non-face related tasks. Rank 64 merges with galaxy and fire risk adapters outperform the baseline.

Age classification adapter can only outperform the baseline when merged with fire and galaxy adapters. However, emotion adapter gives similar performance for rank 16. Interestingly the age regression adapter, which is trained on the same dataset but on a different task, creates a huge performance drop on this task. Lastly the emotion adapter outperforms the baseline in most cases.

## B. How does each adapter effect the others?

In this section we examine the effect of each adapter to the other. Galaxy and fire risk adapters enable other adapters to preserve their performance. In every case their merges with other adapters outperform the baseline. Also, when merged with landmark adapter although their performance drops on their own tasks they preserve performance on the landmark task. They both maintain the performance of the landmark detection task.

Landmark adapter heavily effects other adapters and drops their performance drastically. On all tasks every worst performing merge combination includes landmark adapter. Interestingly it allows fire risk and galaxy adapters to preserve performance up to some level, where as other classification tasks go even below random guess threshold. Landmark adapter creates a huge drop in performance in these face related tasks. It should be noted that the output size of landmark model is much greater than the others since it predicts 98 landmark points represented with 2D coordinates. So, both the output size and the task being a regression task can play role in adapter dominating the others this much.

Age regression adapter doesn't create a huge performance drop as landmark adapter however it drops performance more than the others. Lastly emotion and age classification adapters doesn't have any significant effect on others. Other than some exceptions they preserve performance of the other adapter although not in the level of galaxy or fire risk adapters.

Overall, we observe that every merge decreases the performance in varying levels. Adapters trained on datasets with dissimilar content are resilient to both preserving their own performance and the performance of the other adapter. Adapters trained on similar content tend to lower each other's performance in varying levels, with even going down the random guess threshold.

## C. Going beyond two adapters

To see if more than two adapters can be merged we experimented with merging three models. Since there are too many model combinations we selected fire risk and galaxy classification adapters as the first two adapters and checked their compatibility with other adapters. In Table 8. the performance of three merged adapters are given, where in each case galaxy and fire risk adapters are the two base adapters and the third varies. Only the name of the third adapter is given to prevent confusion and redundancy along with the metric used to evaluate the performance. The results of three adapter merging show that the merged model can still beat the baseline with the exception of landmark adapter as it degrades the performance of fire risk and galaxy adapters. We can see that the overall performance has been slightly degraded compared to merging two adapters. However, as the merged model still beats the baseline merging three models seems achievable.

## IV. DISCUSSION

In this study our goal was to explore if LoRA adapter merging could be used in computer vision tasks to create a multi task model. We trained multiple LoRA adapters on multiple tasks and investigated their performance when they are merged. In our experiments we reported our results by merging up to three models. Our results show that although merged models lose performance it possible to merge multiple LoRAs to create a multitask model. We have found that LoRAs trained on dissimilar content tend to work better with each other. Due to this it is not guaranteed that every merge will preserve performance of the original adapters. Another limitation is that it is more likely for LoRA models with similar content to get merged compared to dissimilar tasks.